\documentclass[letterpaper, 10 pt, conference]{ieeeconf}  

\usepackage{cite}
\usepackage{graphicx}
\usepackage[font=small,labelfont=bf]{caption}
\usepackage[font=small,labelfont=bf]{subcaption}
\usepackage{float}
\usepackage[T1]{fontenc}
\usepackage{multirow}
\usepackage{enumerate}
\usepackage{amsmath}
\usepackage{adjustbox}

\usepackage[dvipsnames]{xcolor}

\definecolor{lightblue}{RGB}{100, 149, 237}

\usepackage{colortbl}
\usepackage{multirow}
\usepackage[para]{footmisc}
\usepackage{lipsum}
\usepackage{booktabs}

\usepackage{xspace}
\usepackage{hyperref}

\makeatletter
\DeclareRobustCommand\onedot{\futurelet\@let@token\@onedot}
\def\@onedot{\ifx\@let@token.\else.\null\fi\xspace}

\def\etal{\emph{et al}\onedot}
\makeatother
\usepackage{amsmath}
\usepackage{amssymb}
\usepackage{dsfont}
\usepackage{etoolbox}
\usepackage{color}
\usepackage{ulem}

\newif\ifshowedits

\newcommand{\addeditor}[3]{%
  \definecolor{#1color}{rgb}{#3}
  \expandafter\newcommand\csname #1\endcsname[1]{%
  \ifshowedits
    {\color{#1color} ##1}%
  \else
    {##1}%
  \fi
  }%
  \expandafter\newcommand\csname #1rmk\endcsname[1]{%
  \ifshowedits
    {\color{#1color} {\bf [#2: ##1]}}
  \fi
  }%
  \expandafter\newcommand\csname #1rpl\endcsname[2]{%
  \ifshowedits
    {\color{#1color} ##1 \sout{##2}}
  \else
    {##1}
  \fi
  }%
}

\newcommand{\createtextvar}[1]{
  \expandafter\newcommand\csname #1\endcsname{%
  {\text{#1}}
}%
}
\newcommand{\mycomment}[1]{}

\newcommand{\vcomment}[1]{}

\addeditor{julien}{JM}{1.0, 0.349, 0.0}
\addeditor{boris}{BM}{0.0, 0.5, 0.0}
\addeditor{mathieu}{MG}{0,0,1}
\showeditstrue

\begin{document}

\title{\LARGE \bf
Leveraging CVAE for Joint Configuration Estimation of Multifingered Grippers from Point Cloud Data
}

\author{Julien Mérand$^{1}$, Boris Meden$^{1}$ and Mathieu Grossard$^{1}$\\ $^{1}$Université Paris-Saclay, CEA, List}

\maketitle
\thispagestyle{empty}
\pagestyle{empty}

\begin{abstract}

This paper presents an efficient approach for determining the joint configuration of a multifingered gripper solely from the point cloud data of its poly-articulated chain, as generated by visual sensors, simulations or even generative neural networks. Well-known inverse kinematics (IK) techniques can provide mathematically exact solutions (when they exist) for joint configuration determination based solely on the fingertip pose, but often require post-hoc decision-making by considering the positions of all intermediate phalanges in the gripper's fingers, or rely on algorithms to numerically approximate solutions for more complex kinematics. In contrast, our method leverages machine learning to implicitly overcome these challenges. 
This is achieved through a Conditional Variational Auto-Encoder (CVAE), which takes point cloud data of key structural elements as input and reconstructs the corresponding joint configurations. We validate our approach on the MultiDex grasping dataset using the Allegro Hand, operating within 0.05 milliseconds and achieving accuracy comparable to state-of-the-art methods. This highlights the effectiveness of our pipeline for joint configuration estimation within the broader context of AI-driven techniques for grasp planning.

\end{abstract}

\section{Introduction}
\label{sec:intro}

Determining joint configurations for multi-fingered robotic grippers is a critical challenge from a control perspective, as precise joint control is essential for accurately positioning fingertips or phalanges at the desired contact points on the object. Indeed, several well-known approaches for generating valid grasps rely on analytical metrics, such as force- or form-closure criteria~\cite{rodriguez2012caging, prattichizzo2012manipulability, rosales2012synthesis}. These methods heavily depend on the knowledge of the reachable contact points to analyze grasp properties~\cite{prattichizzo2016grasping}. From a geometric perspective, the selected grasp configuration must be kinematically reachable by the fingers and collision-free with respect to the environment. This often requires extensive simulation trials to test the accessibility of contact point, implicitly computing the Inverse Kinematics (IK) of each robotic finger.

Although mathematically solvable with an analytical foundation in some cases, the IK problem is not always straightforward, complicating reliability and efficiency in dynamic environments. IK problems can yield multiple solutions — or even an infinite number of solutions in the case of redundant systems — where the poses of intermediate phalanges significantly influence the determination of the appropriate solution. This complexity can also hinder the numerical approximation of solutions. For example, ~\cite{escorcia2023iterative} tackles this challenge by iteratively solving a system of algebraic equations to address the IK problem of an underactuated finger mechanism with coupling.

To mitigate this issues, early works explored the use of feed-forward neural networks with a limited number of joints \cite{koker2004study, duka2014neural}, genetic algorithms \cite{koker2013genetic}, and Neuro-Fuzzy Inference Systems \cite{demby2019study} to solve the IK problem. However, these methods often resulted in large errors and typically returned only a single solution for a given input, which limited their applicability in more complex scenarios. More recent approaches have leveraged generative networks~\cite{ren2020learning, ames2022ikflow, gaebert2024generating} to overcome these limitations, providing the complete solution space and enabling the selection of the most appropriate solution afterwards.

\begin{figure}[t!]
    \centering
    \includegraphics[width=\columnwidth]{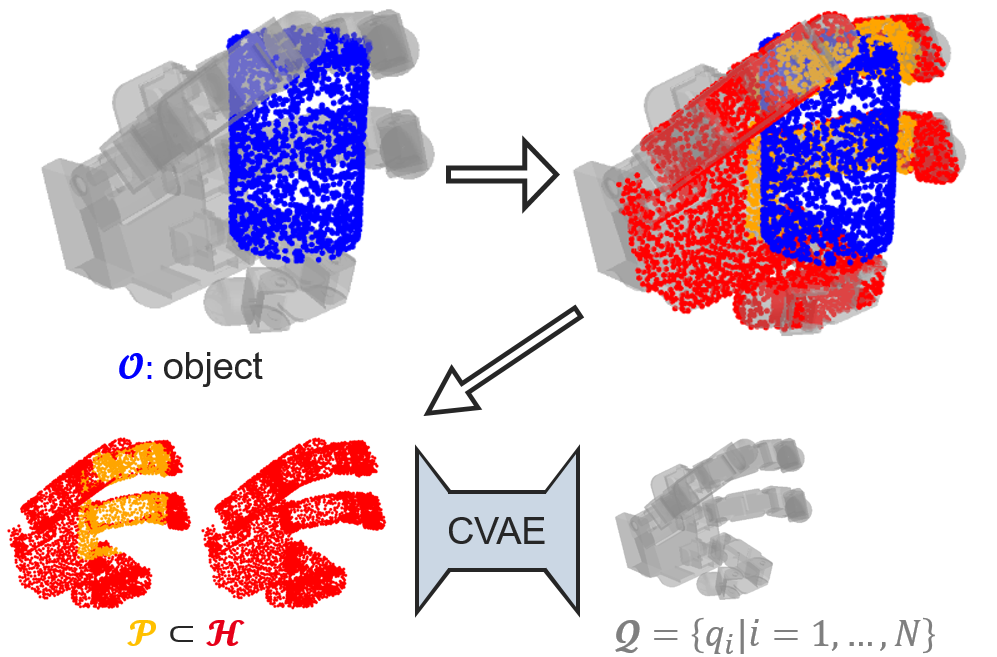}
    \caption{Given a gripper point cloud $\mathcal{H}$ or a set of contact points $\mathcal{P} \subset \mathcal{H}$, our method reconstructs the joint configuration $\mathcal{Q}$. We evaluate this approach in a grasping context using the MultiDex dataset~\cite{li2023gendexgrasp}. 
    }
    \label{fig:teaser}
\end{figure}

Our method builds on this progress by leveraging deep learning techniques to directly infer joint configurations and implicitly select the correct configuration from the set of admissible solutions. This approach aims to streamline the grasping process, enhancing efficiency and adaptability in line with recent state-of-the-art methods that focus on determining the multifingered gripper pose from a 3D Point Cloud (PC) representation of the scene using AI-based paradigms~\cite{newbury2023deep, bohg2013data}. The prevalent approach in the field is to employ numerical optimization processes to derive the optimal joint configuration from this resulting Cartesian fingertip pose~\cite{zhou20176dof, wu2022learning, turpin2022grasp}. While an abundance of models address the initial part of the problem~\cite{shao2020unigrasp, xu2024manifoundation, varley2015generating, wan2023unidexgrasp++, li2023gendexgrasp, wei2024d, lu2024ugg} where data is collected in the form of PC (obtained either through visual sensors or simulations), we specifically focus on an AI-driven approach for determining the joint configuration (Figure~\ref{fig:teaser}). Our aim is to replace the optimization step with an AI-driven approach, aligning with current trends and proposing a method compatible with state-of-the-art practices. By considering intermediate phalanges within the point cloud data, our model recognizes patterns corresponding to specific joint configurations. This eliminates the need for additional decision-making criteria, simplifying the process and enhancing reliability and efficiency in complex manipulation tasks.
 
The paper is organized as follows. In Section~\ref{sec:method}, we introduce the principle of our method, which leverages a Conditional Variational Auto-Encoder (CVAE)~\cite{sohn2015learning} to reconstruct the gripper configuration from its PC representation. We also discuss the relevance of several gripper-related datasets (to account for several practical application cases) and hyperparameters for training, as well as details about the model architecture and its implementation. Section~\ref{sec::evaluation} provides qualitative and quantitative evaluations of our model's performance, using the recent MultiDex grasp dataset~\cite{li2023gendexgrasp} as the core material for evaluation. The influence of the dataset on accurately predicting and reconstructing gripper configurations from new and unseen PCs is also examined. Finally, Section~\ref{sec:conclusion} concludes the paper by summarizing the contributions of our method and offering perspectives for future work.
\section{Method}
\label{sec:method}

\begin{figure*}[!ht]
    \centering
    \includegraphics[width=\textwidth]{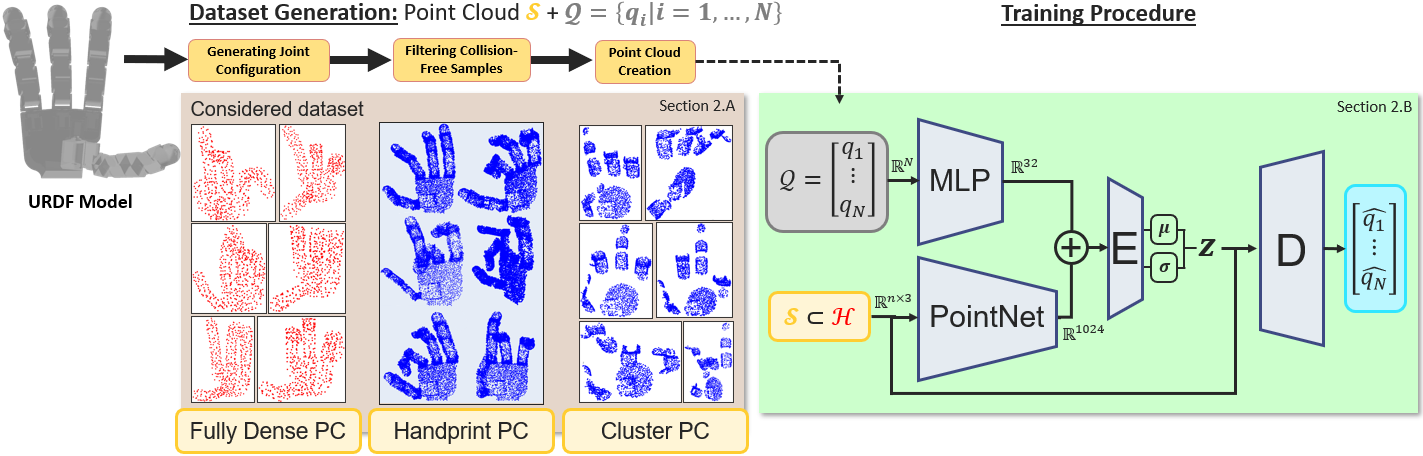}
    \caption{\textbf{Overview of our method:} The left panel illustrates the dataset generation process. Given a gripper URDF model, three datasets of PCs (Fully Dense PC, Handprint PC, Cluster PC) are created as described in Section~\ref{subsec:dataset_generation}, each associated with its joint configuration ($\mathcal{Q}$). 
    The right panel illustrates the model architecture, where a CVAE generates a joint configuration ($\mathcal{Q}$) from a subset ($\mathcal{S}$) of the gripper PC ($\mathcal{H}$). The architecture includes an encoding step that processes the PC using a PointNet, and the joint configuration with a fully connected MLP. These encoded representations are then concatenated and further processed by a latent encoder (E). The decoding step (D) involves processing the encoded PC with a sample from the latent space $z$ (characterized by mean $\mu$ and variance $\sigma$) to reconstruct the joint configuration $\hat{\mathcal{Q}}$ using a fully connected MLP.}
    \label{fig:method_overview}
\end{figure*}

We propose to train a CVAE~\cite{sohn2015learning} to reconstruct a gripper configuration from a point cloud. By training the CVAE on a dataset of PCs paired with corresponding gripper configurations, we aim to develop a model that can accurately predict and reconstruct gripper configurations from new, unseen PCs.

\subsection{Dataset Generation}
\label{subsec:dataset_generation}

Our data generation process is robot-centric and requires only the gripper's URDF to access both its kinematic model and CAD files. The procedure involves the following sequential steps:

\begin{enumerate}
    \item Generating joint configurations: A total of $M$ joint configurations are sampled from the range of motion for each joint, based on a uniform distribution on $[0, 1]$ to guarantee training stability.

    \item Filtering collision-free samples: Given the complex workspace of a multifingered gripper, certain configurations can result in self-collisions, where phalanges overlap or intersect. These configurations are not viable for practical applications. To address this, each generated joint configuration is used as input to the analytical Forward Kinematics model to determine the Cartesian pose of each link. We then implement a collision avoidance function, $f_{\text{self\_collision}}$, inspired by \cite{zhu2021toward}, to detect and filter out configurations where any two links are in collision:
    \begin{equation}
    \centering
    \label{eqn:collision}
        f_{\text{self\_collision}} = \sum_{i=1}^{L} \sum_{j=1}^{L} \mu_{i,j} \cdot \max\Big(\delta_{i,j} - \|\mathbf{k}_i- \mathbf{k}_j\|_{2}, 0\Big),
    \end{equation}
    where:
    \begin{itemize}
        \item $\mathbf{k}_i$ and $\mathbf{k}_j$ are predefined keypoints representing the center of each link's bounding box, given a gripper configuration.
        
        \item $L$ is the total number of links.
        
        \item $\delta_{i,j}$ is a distance threshold that defines the minimum allowable distance between links.
        \item $\mu_{i,j}$ is a weighting factor, set to 1 here.
        
    \end{itemize}
    If $f_{\text{self\_collision}}$ is non-zero, it indicates that at least two links are in collision, and the configuration is deemed invalid.
    
    \item Point Cloud creation: Finally, each link's mesh surface is evenly sampled at varying densities to create a gripper PC.
\end{enumerate}

To demonstrate the model's capability to address a wide range of applications, three distinct datasets are covered following the previous procedure.
\begin{itemize}
    \item \textbf{Fully Dense PC:} This dataset consists of PCs with 512 points equally spaced around the entire gripper. It is designed to benefit methods that map a gripper point cloud to an object, such as in~\cite{wei2024d}.
   
   \item \textbf{Cluster PC:} This dataset uses a small yet representative set of points from each link, aligning with the definition of IK. 
   It ensures that each link is represented by at least one cluster of points, capturing essential spatial information for an accurate determination of the joint configuration. 
   For a given link, a cluster is defined as the set of all points that lie within a sphere of radius $R$. The sphere is centered on the inner surface of the link. This dataset is designed for object-oriented methods, focusing on the regions of the gripper that make contact with the held object through a set of contact points (denoted as $\mathcal{P}$ in Figure~\ref{fig:teaser}).
   
   \item \textbf{Handprint PC:} This dataset focuses on relevant parts of the gripper by filtering surface points using a dot product criterion. It retains only the points that belong to the inner side of the fingers and the palm. Specifically, the dot product is computed between the outward normal of the palm and each point's normal when the hand is in the nominal configuration, with all joints set to 0 radians. This dataset offers a balanced representation, explicitly capturing the gripper's features, especially around the joints, and is closer to real-world measurements, making it a practical compromise between the other two datasets.
\end{itemize}    

All configurations within these datasets were considered within the same reference frame.

To ensure that the entire workspace is covered to achieve maximum generality, we assessed the diversity of all generated joint configurations by calculating the standard deviation of the joint angles' distribution (refer to Section~\ref{subsec:eval_dataset}). This measurement helps verify that the configurations are varied and comprehensive, effectively covering the full range of possible finger motions.

\subsection{Model Architecture}

Given a complete multifingered robotic gripper point cloud, $\mathcal{H} \in \mathbb{R}^{n \times 3}$, we define a subset of points $\mathcal{S} = \{ p \in \mathcal{H} \}$. The overall objective of the model is to retrieve the $N$ joints parameters of the gripper $\mathcal{Q} = \{ q_{i}, i \in [1,N] \}$. \par

The CVAE~\cite{sohn2015learning} is trained to reconstruct $\mathcal{Q}$ knowing $\mathcal{S}$.
It consists of three main components: an encoder, a decoder and a prior distribution. 
The encoder (recognition network) aims to map the input $\mathcal{S}$, conditioned on $\mathcal{Q}$, to a latent distribution $q_{\phi}(z|\mathcal{S}, \mathcal{Q})$. 
We employ a PointNet encoder~\cite{qi2017pointnet} to extract features from the points, which are then concatenated with joint value features obtained using a fully connected Multi-Layer Perceptron (MLP). 
The decoder (generation network), denoted as $p_{\theta}(\mathcal{Q}|\mathcal{S},z)$, focuses on reconstructing the output $\mathcal{Q}$ based on the input $\mathcal{S}$ and the latent variable $z$. 
Concretely, the process is reversed compared to the encoder. We first extract the gripper point features using another PointNet, then concatenate the latent variable $z$ with these per-point features, to generate the joint angles using a fully connected MLP.
Finally, the prior distribution $p_{\theta}(z|x)$ serves as a regularizer, ensuring that the latent space follows this prior distribution. We chose to make the prior distribution independent of the input variable $p_{\theta}(z|x) = p_{\theta}(z) \sim \mathcal{N}(0,1)$, to enhance flexibility in the learning process. \par

We train the generative model by maximizing the log-likelihood of 
$p_{\theta, \phi}(y|x)$, where $\theta$ and $\phi$ are the learnable parameters of the encoder and decoder, respectively. To achieve this, we maximize the evidence lower bound (ELBO) between the estimated posterior distribution and the true posterior distribution~\cite{kingma2013auto}. This is equivalent to solving a minimization problem with the following loss function:

\begin{equation}
\label{eqn:loss_cvae}
    \mathcal{L}_\text{CVAE} = \mathcal{L}_\text{recon} + \beta D_{\text{KL}}(q_{\phi}(z|\mathcal{S}, \mathcal{Q}) \| p_{\theta}(z)).
\end{equation}

The first component $\mathcal{L}_\text{recon}$ is the reconstruction loss, representing the log-likelihood of $p_{\theta}(\mathcal{Q}|\mathcal{S},z)$:

\begin{equation}
\label{eqn:loss_recon}
    \begin{aligned}
        \mathcal{L}_\text{recon} &= -\mathbb{E}_{q_{\phi}(z|\mathcal{S}, \mathcal{Q})}[\log p_{\theta}(\mathcal{Q}|\mathcal{S},z)].
    \end{aligned}
\end{equation}

We approximate this term by a standard RMSE between the ground truth $\mathcal{Q}$ and the prediction $\hat{\mathcal{Q}}$. In other words, it computes the pairwise angular distance: 

\begin{equation}
\label{eqn:loss_mse}
    \begin{aligned}
         \mathcal{L}_\text{recon} &= \mathbb{E}\left[\sqrt{\frac{\sum_{i=1}^{N}{(q_i - \hat{q_i})^2}}{N}}\right].
    \end{aligned}
\end{equation}

The second component, the Kullback-Leibler Divergence ($D_{KL}$), is applied to measure how much the learned distribution $q_{\phi}(z|\mathcal{S}, \mathcal{Q})$ diverges from the prior distribution $p_{\theta}(z)$.

These two terms are weighted by the hyperparameter $\beta$ that can be either set to a fixed value or take the form of a monotonic or cyclical schedule in order to mitigate the known KL vanishing problem~\cite{fu2019cyclical}.

\subsection{Implementation details}
\label{subsec:implementation}

\begin{figure}[h!]
    \centering
    \includegraphics[height=0.5\columnwidth]{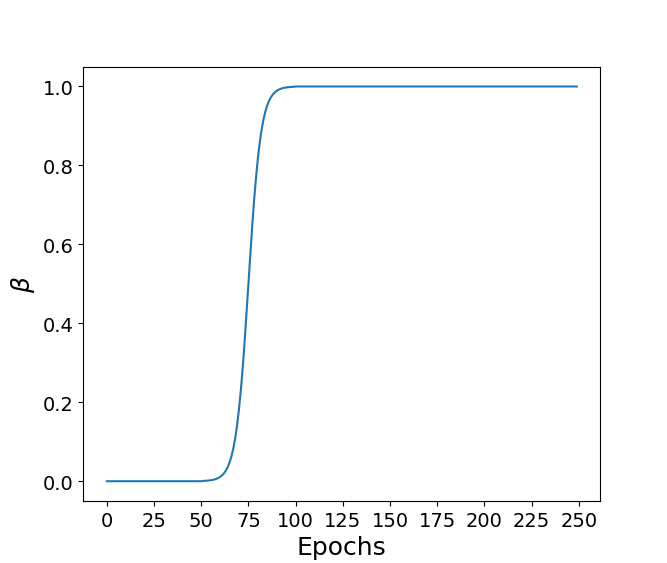}
    \caption{Evolution of $\beta$ over 250 epochs.}
    \label{fig:beta}
\end{figure}

The parameter $\beta$ (see Equation~\ref{eqn:loss_cvae}) was initially set to 0.0001 for the first 50 epochs, then followed a Sigmoid curve ranging from 0.0001 to 1.0 between epochs 50 and 100. After epoch 100, $\beta$ remained constant at 1.0 until the end of training, as we saw no improvements using a cyclical schedule. The evolution of $\beta$ over 250 epochs is illustrated Figure~\ref{fig:beta}.

We defined the dataset size as $M = 30,000$ joint configurations, as it ensures comprehensive coverage of the Allegro's workspace and to facilitate comparison with our evaluation dataset.
$R$ is set to be $50\%$ of each link's radius.
The training process involved 4000 epochs with a batch size of 500, utilizing a single Nvidia GeForce RTX 4090 GPU. Training on the Fully Dense PC dataset required a total of 2 GPU hours. This dataset was split into $80\%$ for training and $20\%$ for testing. We used the Adam optimizer with a learning rate of 0.0001 for optimization.
We conducted our experiment on the Allegro Hand V4~\cite{allegro}, which features $N=16$ independent joints distributed across four fingers—three generic fingers and one thumb—in an anthropomorphic design.

\section{EVALUATION}
\label{sec::evaluation}

Our evaluation of the model's performance leverages the extensive MultiDex dataset~\cite{li2023gendexgrasp}. We start with a qualitative analysis comparing this dataset to our training data. Following this, we conduct a quantitative evaluation to assess the model's accuracy and efficiency using various robotics-based metrics.

\subsection{Dataset: Qualitative analysis}
\label{subsec:eval_dataset}

We evaluated the accuracy of our method using the extensive grasp dataset MultiDex~\cite{li2023gendexgrasp}, created by Li \etal. 
This dataset includes 5 multifingered grippers (EZGripper, Barrett, Robotiq-3F, Allegro, and ShadowHand), 58 household objects from YCB~\cite{calli2015ycb} and ContactDB~\cite{brahmbhatt2019contactdb}, and 436,000 diverse grasping poses. 
We considered the grasping poses of the Allegro Hand, resulting in 37,774 hand PC, compared to the 30,000 PC in our training dataset. The PC are sampled following the same procedure as for our three datasets (Fully Dense PC, Handprint PC, Cluster PC).

Upon examining these configurations, we noticed frequent instances of finger collisions or overlaps. Applying our collision criteria (Equation~\ref{eqn:collision}), only $64\%$ (24,077 out of 37,774) of the configurations would have been retained. However, to demonstrate our model's ability to generalize, we included these configurations in the validation set, meaning that at least $36\%$ of the MultiDex dataset~\cite{li2023gendexgrasp} is not included in our training dataset.

\begin{figure}[!ht]
    \centering
    \includegraphics[width=\columnwidth]{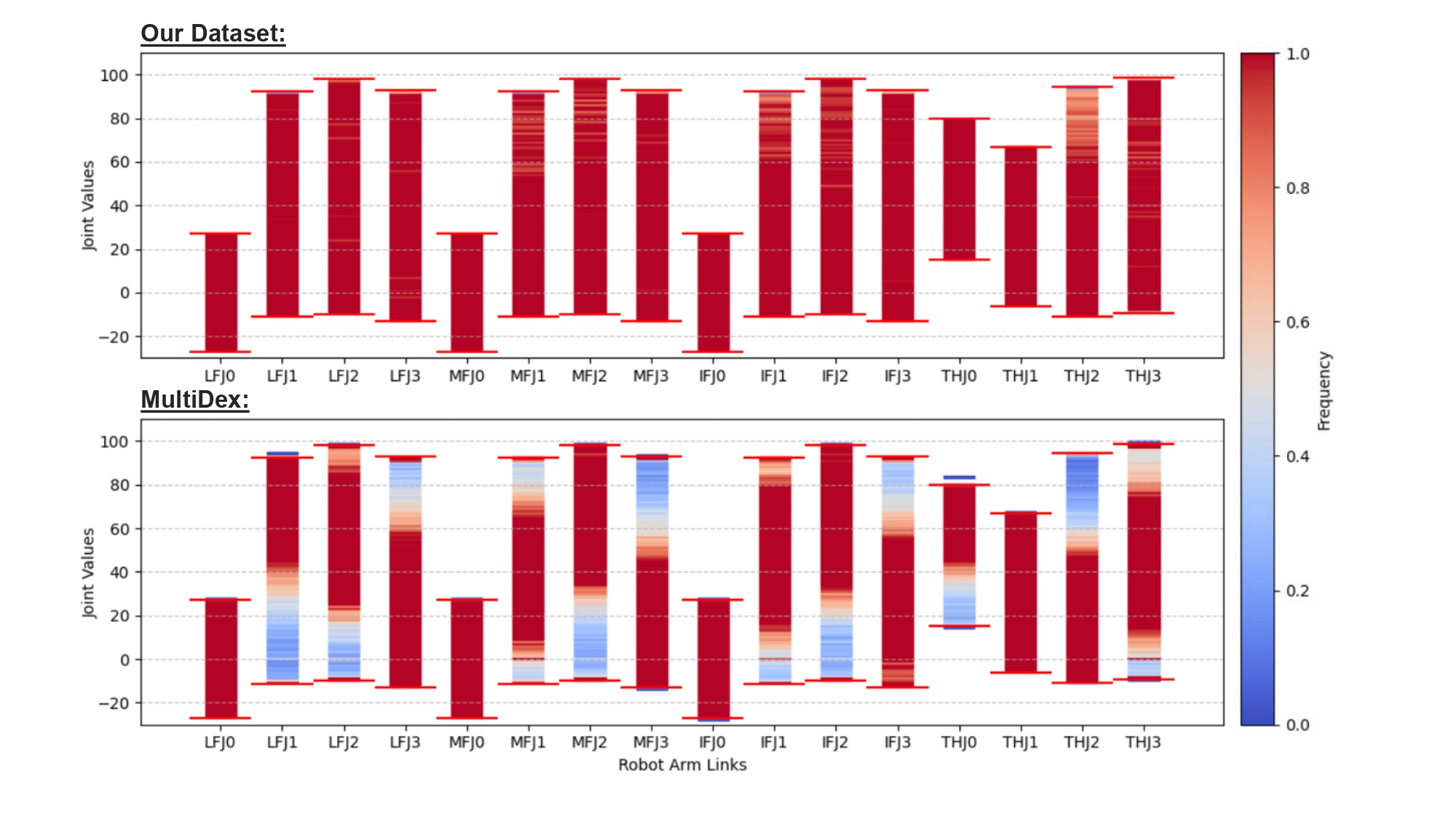}
    \caption{\textbf{Joint Distribution of the Allegro Hand for both datasets.} Top: Fully Dense PC, Bottom: MultiDex dataset~\cite{li2023gendexgrasp}.}
    \label{fig:joint_distribution}
\end{figure}

\begin{figure}[!h]
    \centering
    \begin{subfigure}{\columnwidth}
        \centering
        \includegraphics[width=\columnwidth]{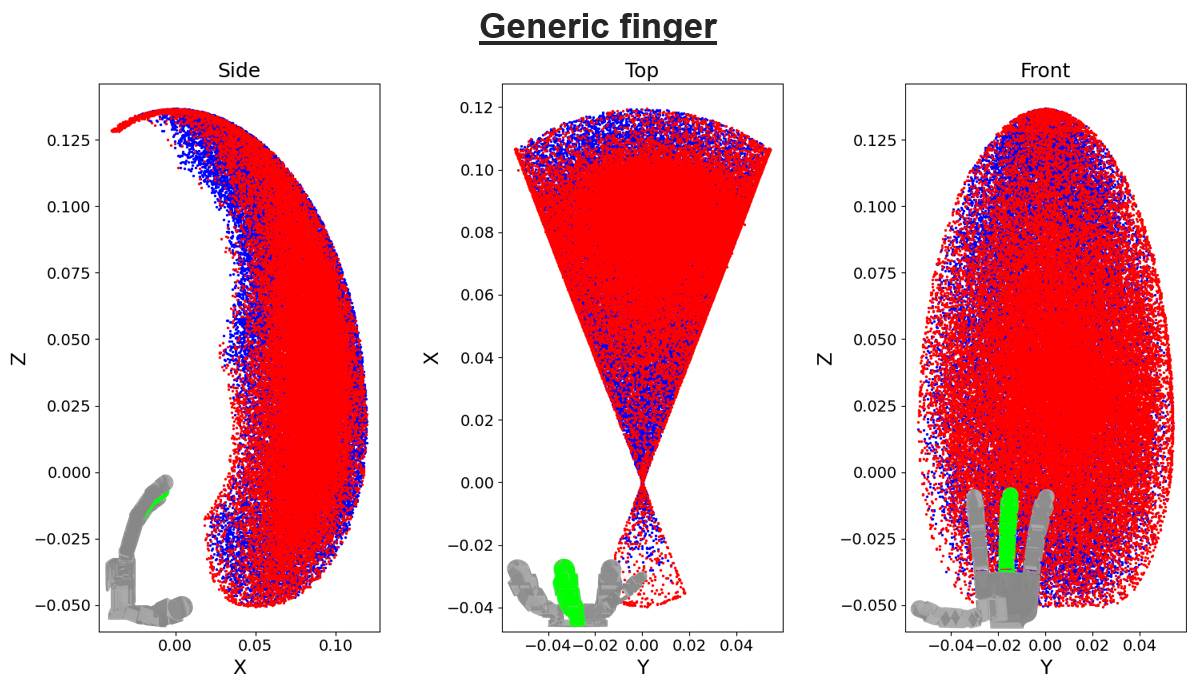}
    \end{subfigure}
    \begin{subfigure}{\columnwidth}
        \centering
        \includegraphics[width=0.3\columnwidth]{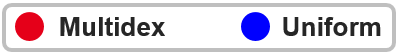}
    \end{subfigure}
    \begin{subfigure}{\columnwidth}
        \centering
        \includegraphics[width=\columnwidth]{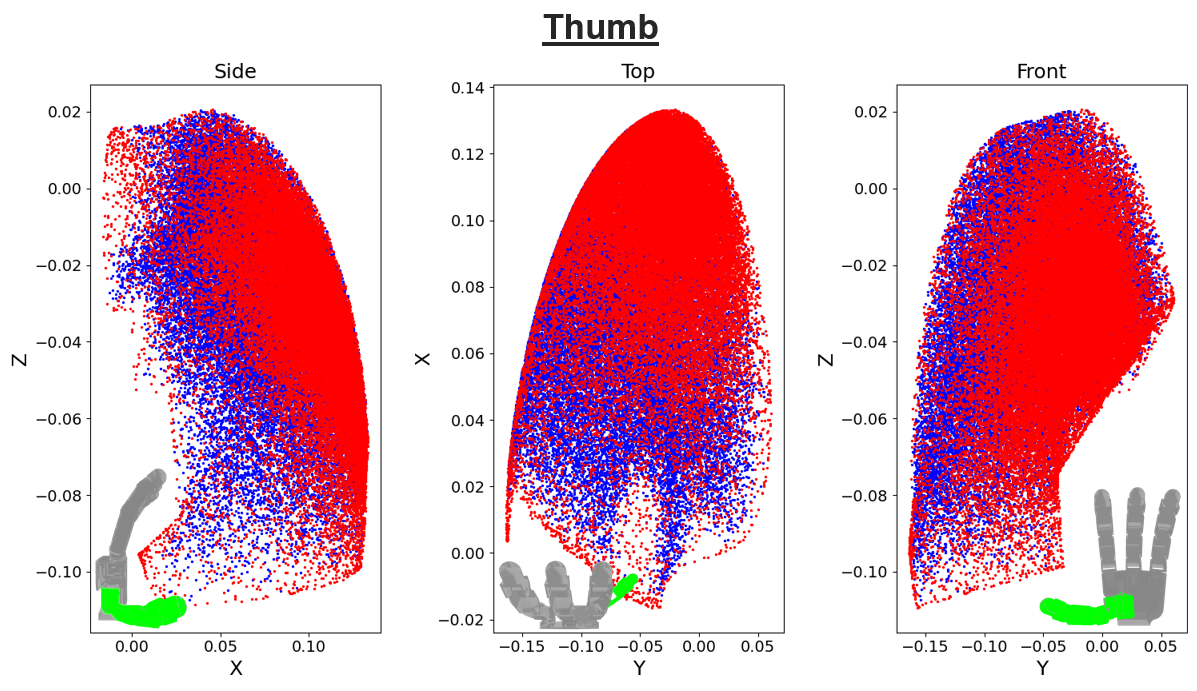}
    \end{subfigure}
    \caption{\textbf{Workspace analysis of the generic and thumb fingers.}}
    \label{fig:workspace}
\end{figure}

Our proposed dataset is uniformly distributed, reflecting a comprehensive coverage of the gripper's workspace (Figure~\ref{fig:workspace}), while the distribution in the MultiDex dataset~\cite{li2023gendexgrasp} is not uniform, likely due to its focus on grasp-related poses.  Our dataset exhibited a standard deviation of 0.287, compared to 0.250 for the MultiDex dataset~\cite{li2023gendexgrasp} (Figure~\ref{fig:joint_distribution}). This aligns closely with the expected standard deviation of a uniform distribution on $[0,1]$, which is $\frac{1}{\sqrt{12}} \approx 0.2887$.

Figure~\ref{fig:workspace} illustrates the workspace of the generic finger and the thumb from a distal perspective, comparing our dataset Fully Dense PC (blue dots) with the MultiDex dataset~\cite{li2023gendexgrasp} (red dots). The workspaces of the generic fingers of the Allegro Hand, along with the thumb, which exhibits a distinct kinematic diagram, are analyzed. This visualization highlights that the MultiDex dataset~\cite{li2023gendexgrasp} does not adequately represent certain regions of the entire workspace. It is worth noting that areas close to the root axis of the thumb are not as well covered in our dataset. This discrepancy is due to our collision avoidance criteria, which do not consider unrealistic configurations from a practical point of view.

\subsection{Quantitative Evaluation}

\begin{table*}[h]
    \begin{center}
    \resizebox{\textwidth}{!}{ 
    \begin{tabular}{l l || c c c | c c}
        \toprule
        \textbf{Metrics} & & \textbf{Fully Dense PC} & \textbf{Cluster PC} & \textbf{Handprint PC} & \textbf{Generic finger} & \textbf{Thumb} \\
        \midrule
        \multirow{2}{*}{\textbf{Mean Joint error}} & \multirow{2}{*}{(rad $\downarrow$)} & $0.075 \pm 0.042$ & $0.064 \pm 0.029$ & $0.063 \pm 0.035$ & $0.017 \pm 0.012$ & $0.018 \pm 0.009$ \\
        & & $4.49\%$ & $3.90\%$ & $3.84\%$ & $1.04\%$ & $1.10\%$ \\
        \midrule
        \textbf{Mean of Lowest Joint error} & \multirow{2}{*}{(rad $\downarrow$)} & $0.022 \pm 0.003$ & $0.029 \pm 0.003$ & $0.022 \pm 0.003$ & $0.002 \pm 0.0005$ & $0.002 \pm 0.0004$ \\
        \textbf{per batch} & & $1.43\%$ & $1.94\%$ & $1.43\%$ & $0.13\%$ & $0.12\%$ \\
        \midrule
        \multirow{2}{*}{\textbf{Mean Cartesian error}} & \multirow{2}{*}{(mm $\downarrow$)} & $3.889 \pm 2.421$ & $3.967 \pm 1.633$ & $3.821 \pm 2.023$ & $0.219 \pm 0.178$ & $0.363 \pm 0.380$ \\
        & & $2.89\%$ & $3.07\%$ & $2.85\%$ & $0.64\%$ & $1.09\%$ \\
        \midrule
        \textbf{Mean of Lowest Cartesian error} & \multirow{2}{*}{(mm $\downarrow$)} & $1.331 \pm 0.176$ & $1.911 \pm 0.227$ & $1.475 \pm 0.223$ & $0.031 \pm 0.009$ & $0.024 \pm 0.006$ \\
        \textbf{per batch} & & $1.03\%$ & $1.54\%$ & $1.13\%$ & $0.09\%$ & $0.07\%$ \\
        \midrule
        \textbf{Time} & (ms $\downarrow$) & $0.042$ & $0.050$ & $0.048$ & $0.044$ & $0.043$ \\
        \bottomrule
    \end{tabular}
    } 
    \end{center}
    \caption{\textbf{Performance evaluation on MultiDex~\cite{li2023gendexgrasp} dataset.} We evaluate our model's ability to estimate gripper configurations using PCs from the MultiDex dataset~\cite{li2023gendexgrasp}. The evaluation considers both Joint error and Cartesian error metrics.}
    \label{tab:MultiDexEval}
\end{table*}

To evaluate the performance of our model, we considered the following quantitative metrics: \par
\begin{itemize}
    \item \textbf{Mean Joint Error (rad)}:  This metric represents the mean error across all joints. An error of 0.01 radians indicates that, on average, each joint in a predicted configuration deviates by 0.01 radians from the ground truth. This metric provides a direct measure of the precision in joint angle predictions.
    \item \textbf{Mean Cartesian error (mm)} : This metric is the mean of the L2 norm between ground truth and predicted keypoints, using the same keypoints as in Equation~\ref{eqn:collision}. It focuses on positional accuracy, as rotational errors are implicitly considered in the finger kinematic diagram. This metric complements joint error by capturing the cumulative effect of errors along the poly-articulated chain. For example, a small error at the finger's base can lead to a larger displacement error at the fingertip.
    \item \textbf{Inference time (ms)}: We measured the inference time on a single Nvidia GeForce RTX 4090 GPU, using a batch size of 100, and computed the mean inference time across all batches.
    This metric is crucial for evaluating the perspective of embedding our model in a robotic  time-constrained controller.
\end{itemize}

Table~\ref{tab:MultiDexEval} presents the results achieved on the MultiDex dataset~\cite{li2023gendexgrasp}. The error rate is relative to the maximum achievable error. For joint error, this corresponds to the full range of each joint. For Cartesian error, it corresponds to the maximal distance between keypoints, determined by evaluating the Euclidean distance between the extreme positions of each keypoint, considering the full range of motion for each joint.

The resulting processing time for inferring the joint configuration is less than $0.05$ ms, using the same resources as those employed for training (Section~\ref{subsec:implementation}). This efficiency is fully compatible with the real-time constraints of robot controllers, which typically operate within a few milliseconds. Therefore, this approach is well-suited for integration into various gripper control strategies.

Comparing our datasets —Fully Dense PC, Handprint PC, and Cluster PC— showed varying levels of accuracy and efficiency. The Handprint PC dataset, for instance, exhibited improved accuracy compared to the Cluster PC dataset, likely due to better feature representation since more points describe the gripper configuration. However, the timing differences were minimal, suggesting that our model's efficiency is robust across different data representations.

The results for the generic finger and the thumb are of comparable magnitude, despite the thumb having a more complex kinematic diagram that allows for distinct behavior, such as opposing the three generic fingers. This demonstrates that our model is independent of any specific kinematic structure, and the method can thus be easily generalized to any multifingered gripper.

The table includes the mean of the minimum error per batch, both Joint and Cartesian, to provide insight into the model's potential accuracy. Given that a CVAE inherently maps one input to multiple possible outputs, inferring with multiple latent samples and selecting the best result (i.e., the one with the lowest error) could improve performance. However, this approach would multiply the inference time, making it less suitable for real-time applications.

The achieved accuracy is comparable (all things considered) to other methods that leverage CVAE to compute the IK model of a robot, as demonstrated in~\cite{gaebert2024generating}.

It is important to note that the PointNet encoder, designed to be invariant to geometric transformations, does not fully achieve this invariance in practice. Introducing rotations and translations to the dataset's PCs would require a different encoder architecture. This new architecture must understand and distinguish both the relationships between points defining the joint configuration and the overall position of the point cloud, representing the gripper's 6D pose.

\section{Conclusion}
\label{sec:conclusion}

This paper proposes a CVAE-based pipeline for joint configuration estimation of complex poly-articulated chains using only their Point Cloud representation. The architecture consists of encoding and decoding steps, incorporating PointNet and MLP networks. The results, considering the Allegro Hand and the grasp patterns included in the MultiDex dataset~\cite{li2023gendexgrasp}, demonstrate that the method meets the precision and inference time requirements for robotic grasping. Its performance is on par with state-of-the-art methods, making it a viable option to be included as part of existing AI-driven grasp planning and control strategies. Additionally, a key advantage of this method is its robot-centric approach. Generating training data is quite straightforward, requiring only the gripper's URDF to access both its kinematic model and CAD files. This simplicity significantly enhances the model's adaptability and efficiency.

To expand the applicability of our pipeline to hand pose estimation and improve its robustness across various grasping scenarios, we plan to estimate the gripper's pose, as an element of $SE(3)$, relative to a desired reference frame. From the perspective of grasping tasks, this step is crucial for establishing a mapping between the object pose and the gripper's pose. Therefore, future work will focus on extending the proposed architecture to estimate the gripper's pose while simultaneously optimizing the model's hyperparameters.

\section*{Acknowledgments}
This project has received funding from the European Union’s Horizon Europe research and innovation program under grant agreement nº 101135708 (JARVIS Project).


\addtolength{\textheight}{-12cm}   


\bibliographystyle{IEEEtran}
\bibliography{refs}

\end{document}